%% file: 0-0_main.tex
\documentclass[10pt, conference]{IEEEtran}
\IEEEoverridecommandlockouts
\usepackage{cite}
\usepackage{amsmath,amssymb,amsfonts}
\usepackage{algorithmic}
\usepackage{graphicx}
\usepackage{textcomp}
\usepackage{xcolor}
\def\BibTeX{{\rm B\kern-.05em{\sc i\kern-.025em b}\kern-.08em
    T\kern-.1667em\lower.7ex\hbox{E}\kern-.125emX}}

\usepackage{booktabs} % For formal tables

\graphicspath{{./figures/}}
\usepackage{subcaption}
\usepackage[ruled]{algorithm2e}
\usepackage{caption}
\usepackage{color}
\usepackage{ctable}
\usepackage{siunitx}
\usepackage{enumitem}
\usepackage{multirow}
\usepackage{balance}

\begin{document}

%\title{Noise Robust Neural Coding for Deep Spiking Neural Networks}
%\title{Noise Robust Deep Spiking Neural Networks}
\title{Noise-Robust Deep Spiking Neural Networks with Temporal Information \\
\thanks{\textsuperscript{*}corresponding author: Sungroh Yoon (sryoon@snu.ac.kr)}
}

\author{\IEEEauthorblockN{Seongsik Park\textsuperscript{1,2}, Dongjin Lee\textsuperscript{1}, Sungroh Yoon\textsuperscript{1,2,3*}}
%\author{\IEEEauthorblockN{Seongsik Park, Seijoon Kim, Byunggook Na, Sungroh Yoon\textsuperscript{*}}
\IEEEauthorblockA{
\textsuperscript{1}Department of Electrical and Computer Engineering, Seoul National University, Seoul 08826, South Korea \\
\textsuperscript{2}Institute of New Media and Communications, Seoul National University, Seoul 08826, South Korea \\
\textsuperscript{3}ASRI and Interdisciplinary Program in Artificial Intelligence, Seoul National University, Seoul 08826, South Korea \\
}
}

\IEEEpubid{978-1-6654-3274-0/21/\$31.00 \textcopyright2021 IEEE}

\maketitle

\begin{abstract}
\input{0-abstract}
\end{abstract}


\section{Introduction}
\input{1-introduction.tex}

\section{Background and Related Work}
\input{2-related_work.tex}

\section{Analysis of Noise on Deep SNNs}

\input{3-anal_noise.tex}

\section{Noise Robust Deep SNNs}

\input{4-noise_robust.tex}

\section{Experimental Results}
\input{5-experimental_results.tex}

%\section{Discussion}
%\input{6-discussion.tex}

\section{Conclusion}

\input{7-conclusion.tex}

% acknowledgments
\section*{Acknowledgments}
\begin{small}

\input{8_acknowledgments.tex}
\end{small}

\balance

\bibliographystyle{IEEEtran}
\bibliography{dac2021}

\end{document}

%% file: 0-abstract.tex
% less than 100 words
Spiking neural networks (SNNs) have emerged as energy-efficient neural networks with temporal information.
SNNs have shown a superior efficiency on neuromorphic devices, but the devices are susceptible to noise, which hinders them from being applied in real-world applications.
Several studies have increased noise robustness, but most of them considered neither deep SNNs nor temporal information.
In this paper, we investigate the effect of noise on deep SNNs with various neural coding methods and present a noise-robust deep SNN with temporal information.
With the proposed methods, we have achieved a deep SNN that is efficient and robust to spike deletion and jitter.

%% long - more than 150 words
%Spiking neural networks (SNNs) have attracted the attention of many researchers for their energy-efficient processing with binary spikes.
%Recently, to fully exploit this event-driven computation, various methods of utilizing temporal information in spike trains have been proposed.
%In addition, SNNs have shown a superior efficiency on neuromorphic devices, which have been expected to lead a new era of hardware for artificial intelligence.
%Nevertheless, noise vulnerability is one of the most urgent issues to be addressed for those devices to be used in real-world applications.
%Several studies have been conducted to increase noise robustness, but most of them considered neither deep SNNs nor temporal information.
%In this paper, we investigated the effect of noise on deep SNNs with various temporal coding.
%In addition, we present a noise-robust deep SNN with temporal information to overcome the aforementioned drawback.
%With the proposed methods, we have achieved a deep SNN that is efficient and robust to spike deletion and jitter.

%% file: 1-introduction.tex
% SNN
Deep learning with deep neural networks (DNNs) has shown remarkable results in various fields.
However, the energy consumption of DNNs has hindered the broad application of deep learning as DNNs deal with more complex tasks.
Spiking neural networks (SNNs) have emerged to address the energy consumption issues of deep learning~\cite{park2019fast, kim2020spiking, park2020t2fsnn}.
SNNs have event-driven computing characteristics with binary spikes, which lead to energy-efficient processing.

Their efficiency and performance are significantly affected by neural coding, which defines how to represent and transfer information between neurons.
There are mainly two types of neural coding, which are rate and temporal coding~\cite{gautrais1998rate}.
For the efficient processing in deep SNNs, temporal coding, including phase~\cite{kim2018deep}, burst~\cite{park2019fast}, and time-to-first-spike (TTFS) coding~\cite{park2020t2fsnn}, has been actively studied.
%더 적은 spike - high efficiency but prone to noise

% noise importance -  % neuromorphic device
SNNs have shown higher energy efficiency on neuromorphic architectures~\cite{merolla2014million,davies2018loihi}.
To improve the efficiency further, lots of studies have been conducted about neuromorphic devices~\cite{roy2019towards, merolla2014million,bouvier2019spiking}.
These devices have been expected to increase the efficiency significantly and lead to a new era of hardware for artificial intelligence.
Despite their promising prospect, there are many obstacles to be resolved to utilize the devices in real-world applications.
One of the most urgent issues is vulnerability to noise in various aspects, such as noisy synaptic weights, unstable integration, and noise in spike trains~\cite{querlioz2011simulation,querlioz2013immunity}. %ref by 동진
Unlike CMOS-based neuromorphic architectures with digital signals, the emerging neuromorphic devices usually process the operations in SNNs with analog signals, such as current and voltage. 
Thus, it is essentially required to ensure noise robustness of the neuromorphic devices.

% noise impact on SNN

% previous work and limitations 
% rate coding
% need to train
% shallow network
%There have been several studies to improve the robustness of SNNs to the several types of noise~\cite{cheng2020lisnn, yu2015spiking, zhang2019fast, zheng2018sparse,chowdhury2020towards}.
A lot of research has improved the robustness of SNNs to several types of noise~\cite{cheng2020lisnn, yu2015spiking, zhang2019fast, zheng2018sparse,chowdhury2020towards}.
%They introduced various methods to increase the noise robustness, but most of them focused on SNNs with rate coding rather than temporal coding, which led to inefficient SNNs.
They have introduced various methods to increase the noise robustness, but most of them focused on SNNs with rate coding.
The effect of noise and their approaches have not been evaluated in the case with other neural coding methods, such as temporal coding, which can improve the efficiency of SNNs.
Furthermore, their methods required to train SNNs with target noise to achieve the noise-robust model.
Even if they showed the effectiveness of these approaches in shallow networks, it has not been verified that their methods can be successfully applied to deep SNNs that are difficult to train.

%Among them, the noise in spike train, including spike jitter and deletion, have a critical impact on the accuracy of SNNs [ref].

\IEEEpubidadjcol

% proposed methods
In this work, to overcome the aforementioned limitations, we analyzed the effect of noise on various neural coding schemes with spike deletion and jitter.
With this investigation, we found that information loss due to the spike deletion had a critical impact on the performance of deep SNNs.
In addition, we revealed that all-or-none activation of TTFS coding was beneficial for deletion-robust deep SNNs that were configured through DNN-to-SNN conversion approaches.
For the jitter noise, we confirmed that TTFS coding was vulnerable to spike shift.

Based on the analysis, we propose noise-robust deep SNNs with temporal coding to achieve both efficiency and robustness.
The proposed methods consist of weight scaling and time-to-average-spike (TTAS) coding.
The weight scaling compensates the information loss effectively, which is caused by the deletion noise, but it is less efficient with all-or-none activation characteristic in TTFS.
To improve the effectiveness of weight scaling, we propose TTAS coding, which utilizes both precise and average spike time.
We are inspired by the \textit{phasic bursting} spike pattern~\cite{izhikevich2004model} for the TTAS coding.
To implement a neuron model that generates the phasic burst spikes, we introduce a simplified integrated-and-fire-or-burst neuron model.
With the TTAS coding, we can improve the effectiveness of weight scaling and cancel out the jitter noise in spike time.
The proposed methods improve the noise robustness without additional training procedures of SNNs, which is a suitable approach to DNN-to-SNN conversion methods.
The contributions of this paper are summarized as follows: 
\begin{itemize}[topsep=0pt,itemsep=0ex,partopsep=1ex,parsep=1ex,leftmargin=*]
\item \textbf{In-depth analysis of spike noise on deep SNNs:} We analyze the impact of spike noise on the performance of deep SNNs with various neural coding methods, including rate, phase, burst, and TTFS coding.
\item \textbf{Noise-robust deep SNN:} We propose a noise-robust deep SNN, which consists of weight scaling and TTAS coding to exploit temporal information under synaptic noise, including spike jitter and deletion.
\end{itemize}

%% file: 2-related_work.tex
\subsection{Spiking Neural Networks}
\input{2_1-snn.tex}
\subsection{Noise on Spiking Neural Networks}
\input{2_2-noise_snn.tex}

%% file: 2_1-snn.tex
% SNN 
SNNs, which are considered the third-generation artificial neural networks~\cite{maass1997networks}, consist of spiking neurons and synaptic weights.
Integrate-and-fire (IF) neurons, which is a widely used type of spiking neurons, integrate inputs into the internal state $u$, which is called membrane potential, as follows:
\begin{equation}
\label{eq:vmem}
    \frac{du_{j}^{l}(t)}{dt} = \sum\nolimits_{i}{w_{ij}^{l} z_{i}^{l}(t) + \eta_{j}^{l}(t)} + b_{j}^{l} \textrm{,}
\end{equation}
where $w$ is a synaptic weight, $z$ is the post-synaptic current (PSC) that is induced by input spike, $\eta$ is a reset function, $b$ is a bias, $i$ ($j$) and $l$ are the indices of pre-synaptic (post-synaptic) neuron and layer, respectively~\cite{gerstner2014neuronal,zhang2020temporal}.
The PSC $z(t)$ is formulated by $(\epsilon * S_{\textrm{in}})(t)$, where $S_{\textrm{in}}$ is an input spike train and $\epsilon$ is a spike response kernel.
The reset function $\eta(t)$ is described as $(\mu * S_{\textrm{out}})(t)$, where $S_{\textrm{out}}$ and $\mu$ are output spike train and reset kernel, respectively.

The spike train $S$, which contains binary spikes, is stated as
\begin{equation}
\label{eq:spike_train}
    S_{i}^{l}(t) = \sum\nolimits_{t_{i,f}^{l} \in F_{i}^{l}}{\delta(t-t_{i,f}^{l})} \textrm{,}
\end{equation}
where $\delta$ is the Dirac delta function and $f$ is the index of spike in a set of spikes $F$ satisfying the firing condition as follows:
\begin{equation}
\label{eq:firing_time}
    t_{i,f}^{l}: u_{i}^{l}(t_{i,f}^{l}) \geq \theta_{i}^{l}(t_{i,f}^{l}) \textrm{,}
\end{equation}
where $\theta$ is a threshold.
Due to the features of integrate-and-fire and event-driven computing with binary spike trains (Eqs.~\ref{eq:vmem} and \ref{eq:spike_train}), SNNs have the potential for energy-efficient processing.

% the importance of temporal information in SNNs
SNNs transmit information between neurons with binary spike trains.
Hence, the efficiency and performance of SNNs vary depending on neural coding methods, which define how to represent the information in the form of a spike train.
The neural coding schemes are mainly divided into the rate and temporal coding~\cite{gautrais1998rate}, as shown in Fig.~\ref{fig:nc_noise}-A.
Rate coding utilizes spike firing rate $r=N/T$ to represent the information in a given time window $T$ with the number of spikes $N$~\cite{rueckauer2018conversion, han2020rmp, kim2020spiking}.
This approach is known to be less efficient because it does not exploit temporal information in spike trains~\cite{gautrais1998rate}.
An empirical threshold balancing improved the efficiency, but it did not overcome the drawback of rate coding~\cite{han2020rmp}.

Temporal coding is represented by phase~\cite{kim2018deep}, burst~\cite{park2019fast}, and TTFS coding~\cite{park2020t2fsnn}, which utilize global oscillator, inter-spike interval (ISI), and spike time, respectively.
Among them, TTFS coding uses the least number of spikes, which results in the highest computational efficiency.
The TTFS coding had a latency issue, but it was resolved in \cite{park2020t2fsnn}.
Thus, this neural coding is promising for efficient deep SNNs.

%% file: 2_2-noise_snn.tex
\begin{figure}[t]
    \centering
    \includegraphics[width=1.0\linewidth]{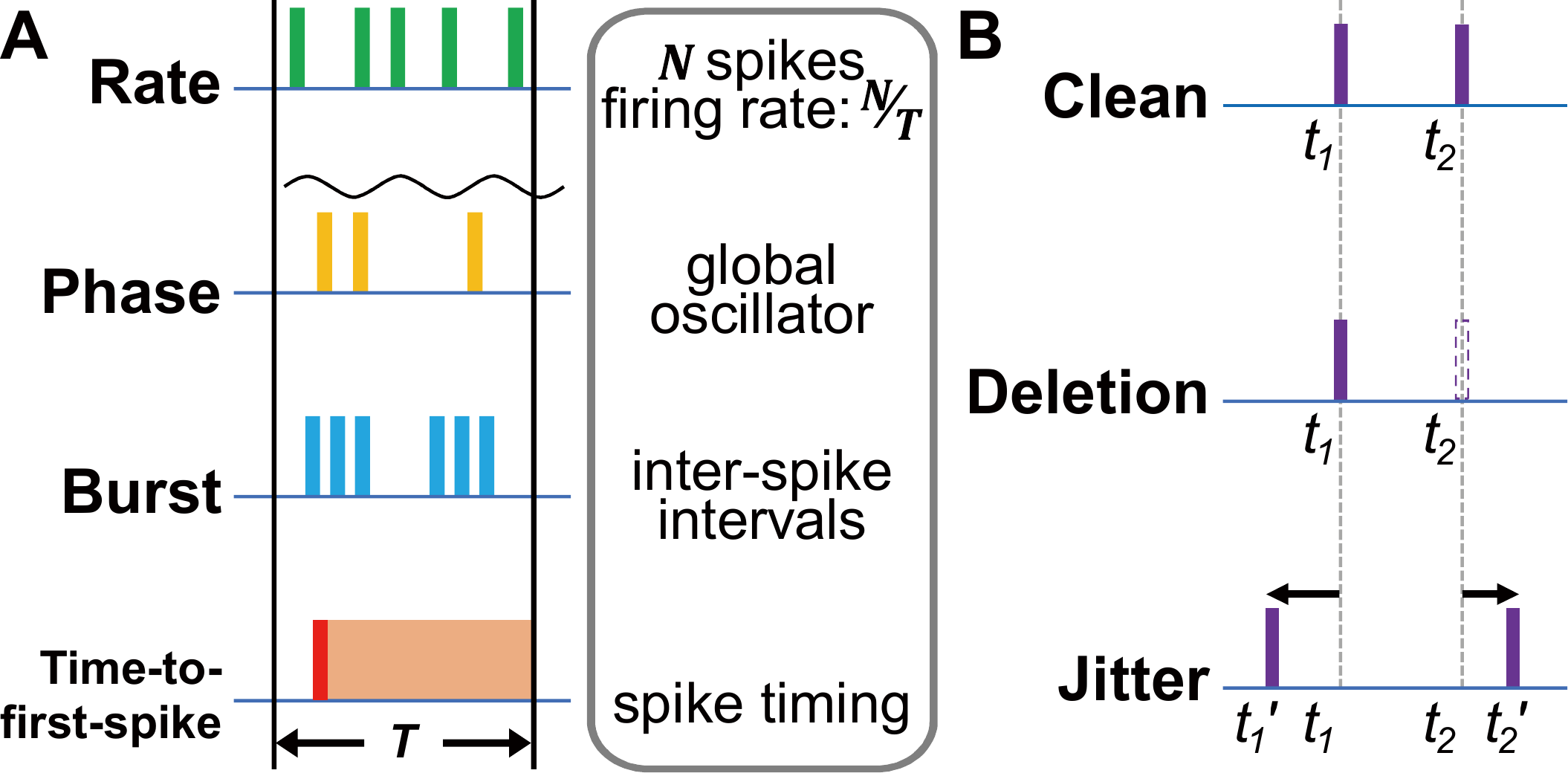}
	\caption{A) various neural coding methods and B) spike noise}
	\label{fig:nc_noise}
	\vspace{0.5em}
\end{figure}

%
%Emerging neuromorphic devices, which are the target platforms of SNNs, have inherent noise.[refs]
%Hence, analyzing the effect of noise on SNNs is an essential task for future implementation.
%Unlike existing Von Neumann architecture based hardware actively deployed in the field of deep learning, such as GPUs, neuromorphic hardware is exposed to various noises since computation is performed through analog circuits.[ref] %intro에도 있으니 제외?

The types of noise associated with neuromorphic hardware can be classified mainly into external and internal noise.
External noise, or input noise, originates from corruption or perturbation of the input data~\cite{hendrycks2019benchmarking}.
The reason for this kind of noise is not directly related to neuromorphic hardware itself.
%However, considering the potential usage of neuromorphic hardware, managing input noise cannot be underestimated since numerous real-world data, which is gathered with sensors, is noisy.
However, managing the input noise cannot be underestimated considering the potential usage of neuromorphic hardware with noisy data in the real world.
Several works have studied the effect of this kind of noise on SNNs~\cite{cheng2020lisnn, yu2015spiking, zhang2019fast, zheng2018sparse,chowdhury2020towards}. % too much?

%From a hardware perspective, internal noise can be further classified as follows.
The internal noise can be further classified as follows: static and dynamic noise.
%The first noise is static noise (termed `fixed-pattern noise' in~\cite{wunderlich2019demonstrating}), which is caused by manufacturing variations of neuromorphic hardware.
The static noise (termed `fixed-pattern noise' in~\cite{wunderlich2019demonstrating}) is caused by manufacturing variations of neuromorphic hardware.
This mismatch leads to parametric errors invariant over time.
%The other noise, dynamic noise (termed `temporal variability' in~\cite{wunderlich2019demonstrating}) is considered more arduous to handle.
The dynamic noise (termed `temporal variability' in~\cite{wunderlich2019demonstrating}) is considered more arduous to handle.
%Mainly caused by thermal variations or instabilities of analog circuits, this noise can cause time-varying deviations in neuromorphic devices.
This noise can cause time-varying deviations in neuromorphic devices since it is mainly caused by thermal variations or instabilities of analog circuits.
Thus, unlike static noise, circuits with dynamic noise can produce different outcomes in multiple trials even when the input remains unchanged.

%From an algorithmic point of view, in order to simulate the noisy nature of neuromorphic hardware aforementioned, SNN neurons can be modeled with the following noises.
%This variations on model parameters include variations on weights, thresholds, or time constants.
%These variations on model parameters include variations on weights, thresholds, or time constants.
%These variations on model parameters were related to weights, thresholds, or time constants.
%\cite{stromatias2015robustness, li2020robustness, neftci2016stochastic} studied the effect of this kind of noise on SNNs.
In order to simulate the aforementioned noisy nature of neuromorphic hardware, SNNs can be modeled with the noise on various parameters, such as synaptic weights, thresholds, and time constants~\cite{stromatias2015robustness, li2020robustness, neftci2016stochastic}.
In another aspect, one can model neurons to produce noisy output spikes, which typically appear in the form of spike jitter and deletion~\cite{yu2020synaptic, wu2018spiking, yu2020robust}.
%There exist a number of previous works that handle these types of noise using various schemes~\cite{yu2020synaptic, wu2018spiking, yu2020robust}.
%Note that noisy spikes are also the by-product of calculation with erroneous model parameters.
Since the parametric errors in the former approach result in noisy spikes, we adopted the latter method for modeling the noise.

%After implementing the network on neuromorphic hardware, errors resulting from the static noise can be mitigated by parameter correction via on-chip learning or 
After implementing SNNs on neuromorphic hardware, errors resulting from the static noise can be mitigated by parameter correction via on-chip learning~\cite{wunderlich2019demonstrating}.
However, errors due to the dynamic noise cannot be resolved in this way.
%Instead, the network needs to be \textit{designed} robust to noise during training, before deployment to the neuromorphic hardware.
%Instead, the network needs to be ``designed'' robust to noise in the first place, before actual implementation to target neuromorphic hardware.
%Instead, the network needs to be \textit{designed} robust to noise in the first place, before actual implementation to target neuromorphic hardware.
%Thus, addressing the noisy spikes issue ought to be the goal for designing noise robust SNNs.
%Thus, to address the noise issues, SNNs need to be \textit{designed} robust to the noise before deployment to the neuromorphic hardware.
Thus, to address the noise issues, we need to design SNNs to be robust to the noise before deployment to the neuromorphic hardware.

%% file: 3-anal_noise.tex
In this study, we analyzed the effect of the spike noises, which are spike deletion and jitter, on deep SNNs with various neural coding methods.
Based on this investigation, we propose a noise-robust deep SNN with temporal information to achieve both high efficiency and noise robustness.
To utilize various neural coding approaches in deep SNNs, we adopted DNN-to-SNN conversion methods as in \cite{rueckauer2018conversion, han2020rmp, kim2018deep, park2019fast, park2020t2fsnn}.
We implemented the spike deletion with a deletion probability $p$ and uniformly distributed random variable between zero and one.
To implement the spike jitter, we used Gaussian noise with zero mean and standard deviation $\sigma$.
The jitter noise was determined by $\sigma$ and quantized into an integer to add it to spike time $t_{f}$.
We swept the deletion probability $p$ from 0.1 to 0.9 and the standard deviation $\sigma$ from 0.5 to 4.0.
To evaluate the effect of noise on deep SNNs, we experimented with the CIFAR-10 dataset on VGG16.

\begin{figure}[t]
    \centering
    \includegraphics[width=\linewidth]{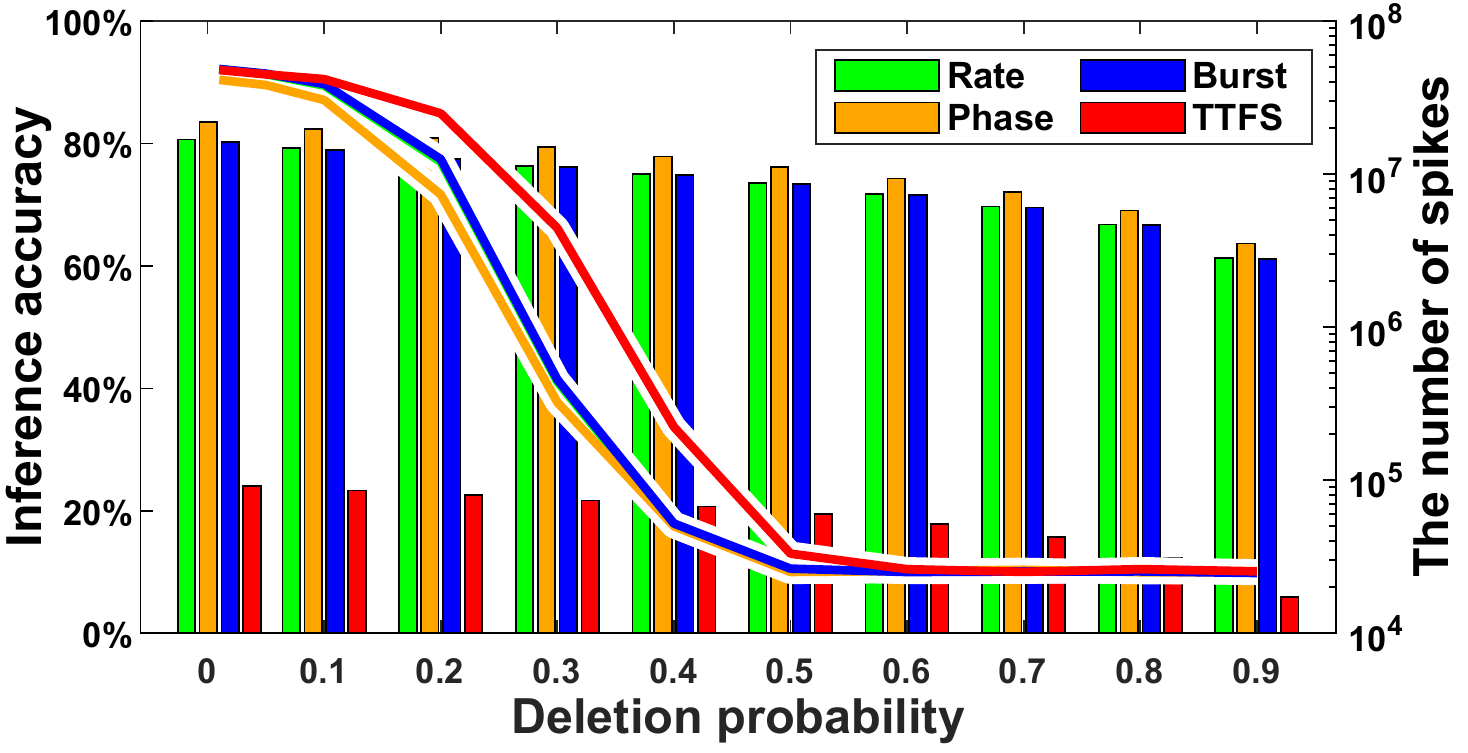}
	%\vspace{-0.5em}
	\caption{Inference accuracy and the number of spikes with spike deletion on VGG16 and CIFAR-10 dataset depending on various neural coding methods}
	\label{fig:nc_del_acc_pr}
	\vspace{-0.5em}
\end{figure}

%analysis of the effect of noise on deep SNNs - jitter and deletion
%how to implement deletion and jitter
%deletion - uniform distribution, sweep deletion probability 1\% to 90\%
%jitter - gaussian probability mean zero, sweep std
%setup - input real value and the first convolution layer - spike generator as \cite{rueckauer2018conversion, han2020rmp}

%The overall accuracy degradation according to the deletion noise is 
%- overall degradation - information loss
%- TTFS robustness - dropout

% deletion 
The results of deletion noise are depicted in Fig.~\ref{fig:nc_del_acc_pr}.
As the probability of deletion increased, the accuracy and the number of spikes decreased in all cases.
When the deletion probability $p$ was greater than 0.4, the accuracy was less than 40\%.
Due to information loss that was caused by the spike deletion, the overall accuracy deteriorated.
With the deletion probability $p$, the sum of PSC $Z'$ during time window $T$, which corresponds to activation $A$ in DNNs, is reduced to $(1-p)Z$ on average, where $Z$ is the sum of PSC without deletion noise.

Among the neural coding methods, TTFS was the most robust to the deletion noise.
The reasons lay in information transmission methods of neural coding and the DNN-to-SNN conversion approach.
Even if the expected activation values of various neural coding methods with a deletion probability are the same with $(1-p)A$, where $A$ is the activation without noise, the effect of the noise on the activation appeared differently.
In rate, phase, and burst coding, a number of spikes carry an activation $A$, which is the sum of PSC.
When the number of spikes is $N$, the effective PSC of a spike is $A/N$.
Hence, each activation with the deletion noise $A'$ lies in a range of $[0, A]$.
On the other hand, in TTFS coding, one spike carries the activation $A$; thus, the deletion-noisy activation $A'$ is $0$ or $A$.

%smaller effective PSC induces fraction of information, which results in larger residual membrane potential

These differences in the noisy activation cause the different noise robustness on various neural coding with the DNN-to-SNN conversion method.
In the conversion approach, which is widely used for training deep SNNs indirectly, synaptic weights of SNNs are converted from pre-trained DNNs; thus, the weights have characteristics resulting from the training of DNNs.
Dropout~\cite{srivastava2014dropout}, which is used for avoiding overfitting and increasing generalization performance of DNNs, affects the noise robustness of each neural coding.
It randomly makes activations zero during training DNNs, which is similar to the deletion noise in TTFS.
This all-or-none property of activation strengthened the robustness to deletion noise of TTFS coding, as shown in Fig.~\ref{fig:nc_del_acc_pr}.

\begin{figure}[t]
    \centering
    \includegraphics[width=1.0\linewidth]{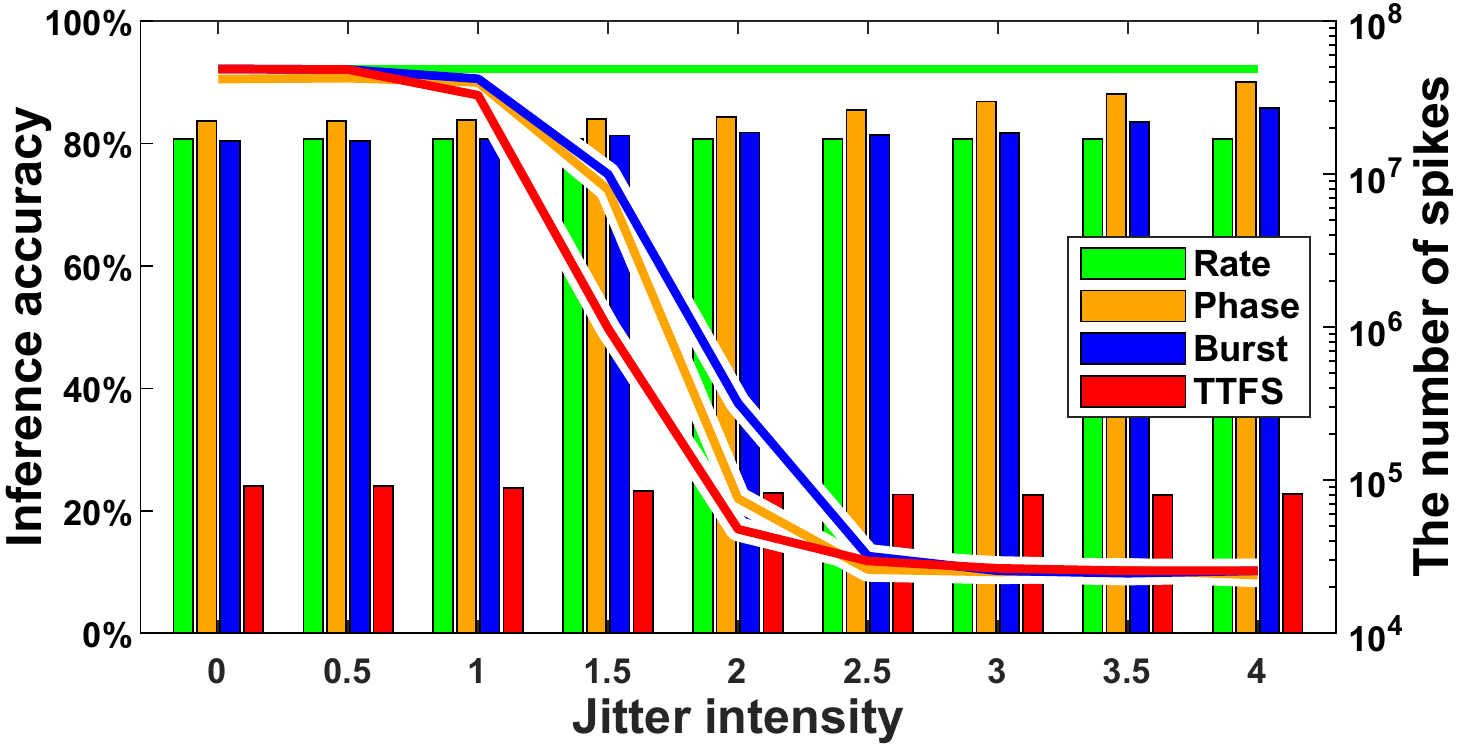}
	%\vspace{-0.5em}
	\caption{Inference accuracy and the number of spikes with spike jitter on VGG16 and CIFAR-10 dataset depending on various neural coding methods}
	\label{fig:nc_jit_acc_pr}
	\vspace{-0.5em}
\end{figure}

% jitter
The experimental results of spike jitter are depicted in Fig.~\ref{fig:nc_jit_acc_pr}.
Rate coding was hardly affected by the spike jitter because it does not utilize temporal information in spike trains.
The jitter noise significantly affected the results of temporal coding methods.
Especially, TTFS was the most susceptible to the spike jitter because it uses only one spike per activation.
The number of spikes on each neural coding did not vary considerably according to the jitter intensity.
Phase and burst coding methods showed a tendency to increase in the number of spikes as jitter intensity increased.
%did not vary according to the neural coding, but phase and burst coding methods showed a tendency of increase in the number of spikes as jitter intensity increased.
In contrast, TTFS presented a much smaller number of spikes, which left room for redundant spikes to cancel out the jitter noise.

%\textcolor{red}{spike 수와 jitter noise robustness 관계}
%TTFS uses one spike per activation, thus it is the most susceptible to the spike jitter.
%Thus, multiple spike를 활용하여 jitter noise를 상쇄할 수 있는 방법이 요구된다.

%rate coding은 spike train 의 temporal information이 아닌 spike firing rate로 정보를 전달하기 때문에, jitter에 대한 영향이 거의 없다.
%Phase coding 은~
%burst coding 은~
%TTFS는 only a spike를 사용하여 정보를 전달하기 때문에 jitter에 가장 큰 영향을 받는다.
%Thus, multiple spike를 활용하여 jitter noise를 상쇄할 수 있는 방법이 요구된다.

%% file: 4-noise_robust.tex
% need to duplicate the information (needs for redundant spikes)
% noise robust 하기 위해서는 정보를 중복해서 전달해야함 
% duplicated spikes의 문제점 on various neural coding methods

% intuition from the analysis
In the previous section, we investigated the vulnerability of temporal coding to spike deletion and jitter.
Despite the high efficiency of TTFS coding in terms of the number of spikes and inference time, the information loss and spike shifting, which were caused by the spike deletion and jitter, respectively, hindered the use of temporal information in deep SNNs in noisy environments, such as emerging neuromorphic devices.
To overcome such obstacles, we propose a noise-robust deep SNN, which consists of weight scaling and TTAS coding.

\begin{figure}[t]
    \centering
    \includegraphics[width=1.0\linewidth]{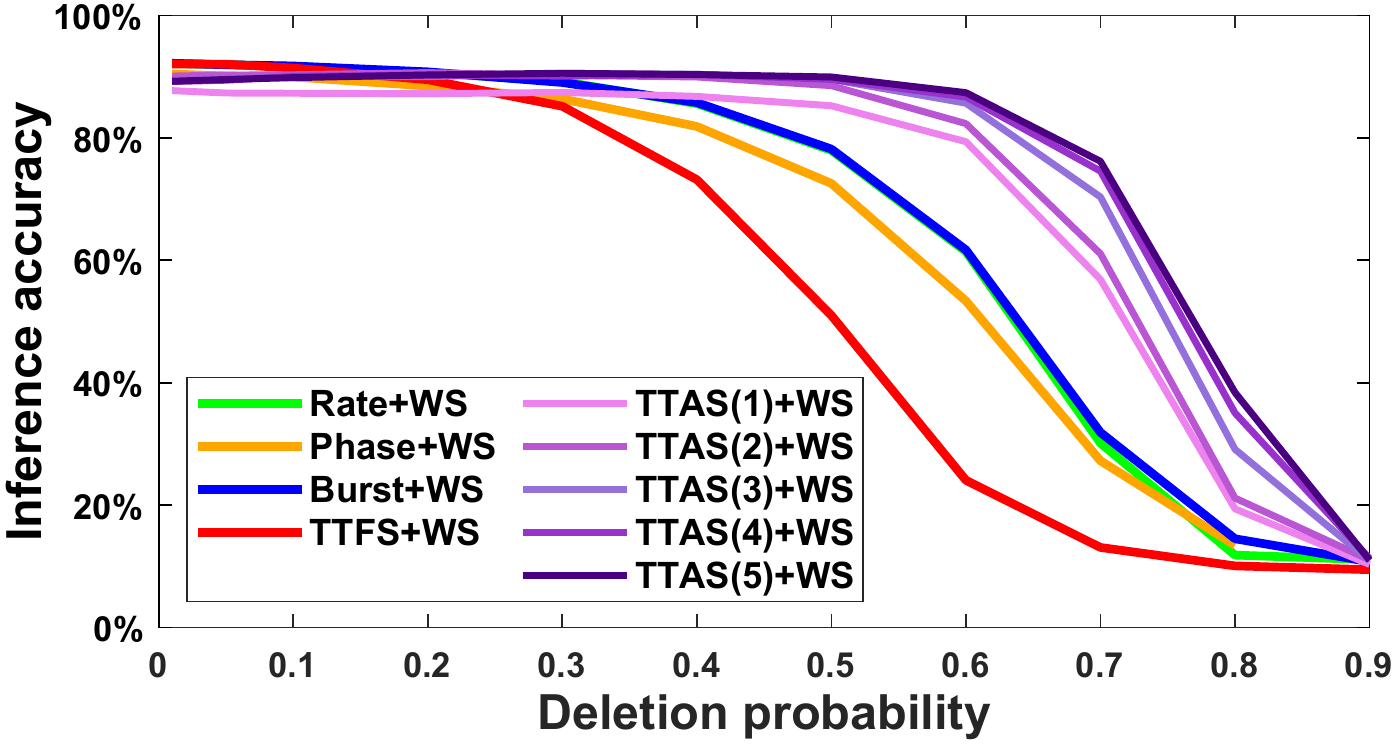}
	%\vspace{-0.5em}
	\caption{Inference accuracy of (weight scaling (WS) and TTAS) with spike deletion on VGG16 and CIFAR-10 dataset depending on various neural coding methods}
	\label{fig:proposed_del}
	%\vspace{-2.0em}
\end{figure}

The weight scaling compensates for the information loss due to the deletion noise.
As discussed in the previous section, the amount of activation is reduced to $A'=(1-p)A$ on average with the deletion probability $p$.
To guarantee sufficient information with the deletion, we scale the synaptic weight as $W' = CW$, where $C$ is a scale factor, and $W'$ is the scaled weight.
As a pioneering study of the weight scaling for the noise robustness, we set $C$ proportional to the deletion probability $p$.
The results of weight scaling are shown in Fig.~\ref{fig:proposed_del}.
The weight scaling approach increased the noise robustness in all cases.
%We can see the improvement by the scaling in differences between dotted and solid lines in each color.

%
Although the robustness of deep SNNs was improved by the scaling, it was less effective in TTFS coding.
The difference in the effectiveness of the scaling originated from the information transmission methods.
In rate, phase, and burst coding, a number of spikes were required to represent an activation, which led to uniformly reduced activation $(1-p)A$ with high probability.
These reduced activations were restored with the deterministic weight scaling factor $C$.
On the other hand, in TTFS coding, the weight scaling resulted in an activation value of $0$ or $CA$ with the probability of $p$ or $(1-p)$, respectively.
Thus, it caused over activations, which led to the least improvement in deletion robustness.

\begin{figure}[t]
    \centering
    \includegraphics[width=1.0\linewidth]{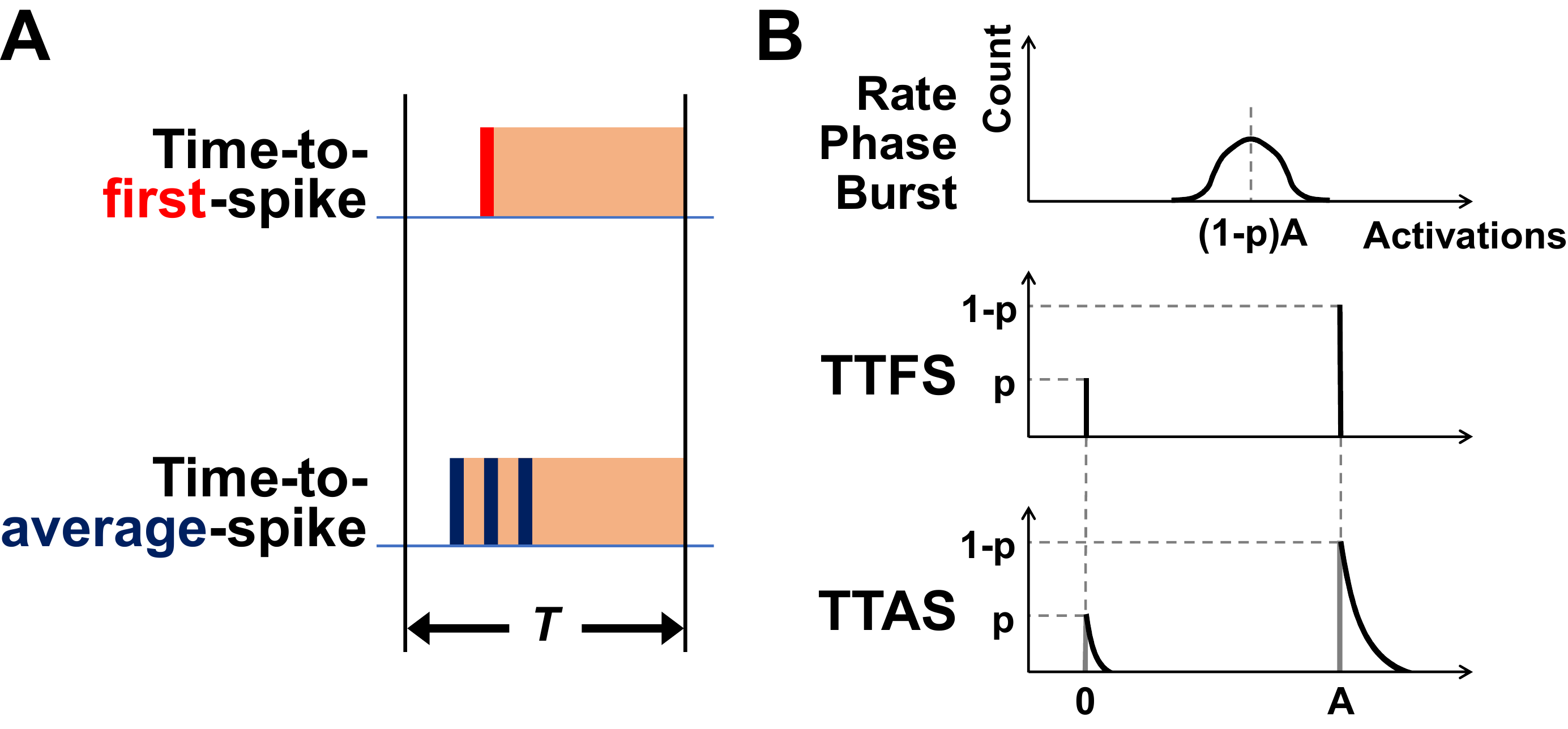}
	%\vspace{-0.5em}
	\caption{A) Comparison between time-to-first-spike (TTFS) and time-to-average-spike (TTAS) and B) activation distribution on various neural coding methods}
	\label{fig:TTFS_and_TTAS}
	\vspace{-0.5em}
\end{figure}

To alleviate the over activation and exploit all-or-none activation property for deletion noise, we propose TTAS coding, which utilizes both precise spike time and burst spikes, as shown in Fig.~\ref{fig:TTFS_and_TTAS}-A.
This temporal coding is inspired by \textit{phasic bursting} spike pattern~\cite{izhikevich2004model}.
To implement the phasic burst spikes with negligible computational overhead, we introduce a new neuron model with a reset function as
\begin{equation}
\label{eq:neuron_model}
    \eta(t) = 
    \begin{cases}
        0               & \textrm{if}~(t < t_{1}) \\
        \theta(t)       & \textrm{if}~(t \geq t_{1}) \operatorname{and} (t < t_{1}+t_{\textrm{a}})\\
        -\infty         & \textrm{otherwise} {,}
    \end{cases}
\end{equation}
where $t_{1}$ and $t_{\textrm{a}}$ are the first spike time and target duration of burst spikes, which indicates the number of phasic burst spikes, respectively.
This neuron model can be considered a simplified integrate-and-fire-or-burst neuron~\cite{wilson1999simplified} and implemented with counter and gate operations.

% additional scaling factor
TTAS coding utilizes redundant spikes in the first group of spikes to deliver information, which increases the sum of PSC as
\begin{equation}
\label{eq:ttas_psc}
    \hat{Z}=\sum_{t}^{t_{\textrm{a}}}{z(t^{1}+t)}\textrm{.}
\end{equation}
For accurate processing, we need to offset this increment.
We set another scaling factor for TTAS coding as $C_{\textrm{A}}=z(t^{1})/\hat{Z}$ and integrated the scale factor to synaptic weight in order not to increase the computational overhead.

% more detailed explanation with figures
Rate, phase, and burst coding methods were vulnerable to deletion noise due to their continuously reduced activation, as depicted in Fig.~\ref{fig:TTFS_and_TTAS}-B.
On the contrary, TTFS was prone to jitter noise because of the all-or-none activation.
Each of them has its advantage; thus, it is crucial to combine the two advantages to achieve efficiency and robustness simultaneously.
With the proposed TTAS coding, we can exploit the benefit of the all-or-none activation feature and information compensation by weight scaling.
In particular, when TTAS is applied to the exponentially decreasing PSC kernel, as in \cite{park2020t2fsnn}, the activation distribution appears high around $0$ and $A$.
%This makes it possible to utilize both the nature of TTFS, which expresses activation discontinuously, and the rate and burst coding, which present continuous representation of the activation with deletion noise.
This distribution makes it possible to exploit both advantages of discontinuous and continuous activations for the deletion and jitter noise.

%
%weight scaling - compensate information loss due to spike noise
%time-to-average spike coding
%Burst spike - duplication or redundant spike
%- phasic burst spikes 
%- TTAS

%noise robust SNN with temporal information
%- phasic burst spikes
%- neuroscience inspired
%- integrate-and-burst model
%- but it induces computational overhead
%- we propose simplified integrate-and-burst neuron model that generates phasic burst spikes

%the reason to choose TTFS as base neural coding
%- rate - duplication (redundant) information (or spikes) causes saturation of firing rate
%- phase - 

%methods
%to minimize the effect of noise in both spike deletion and jitter, we duplicated 

%we propose spiking neuron model and neural coding
%name? - modify reset function  $\eta$
%propose time-to-average-spike (TTAS) coding

%% file: 5-experimental_results.tex
%\subsection{Evaluation of Proposed Methods}

%\subsection{Comparisons with Other Neural Coding Methods}
%

%
\begin{figure}[t]
    \centering
    \includegraphics[width=1.0\linewidth]{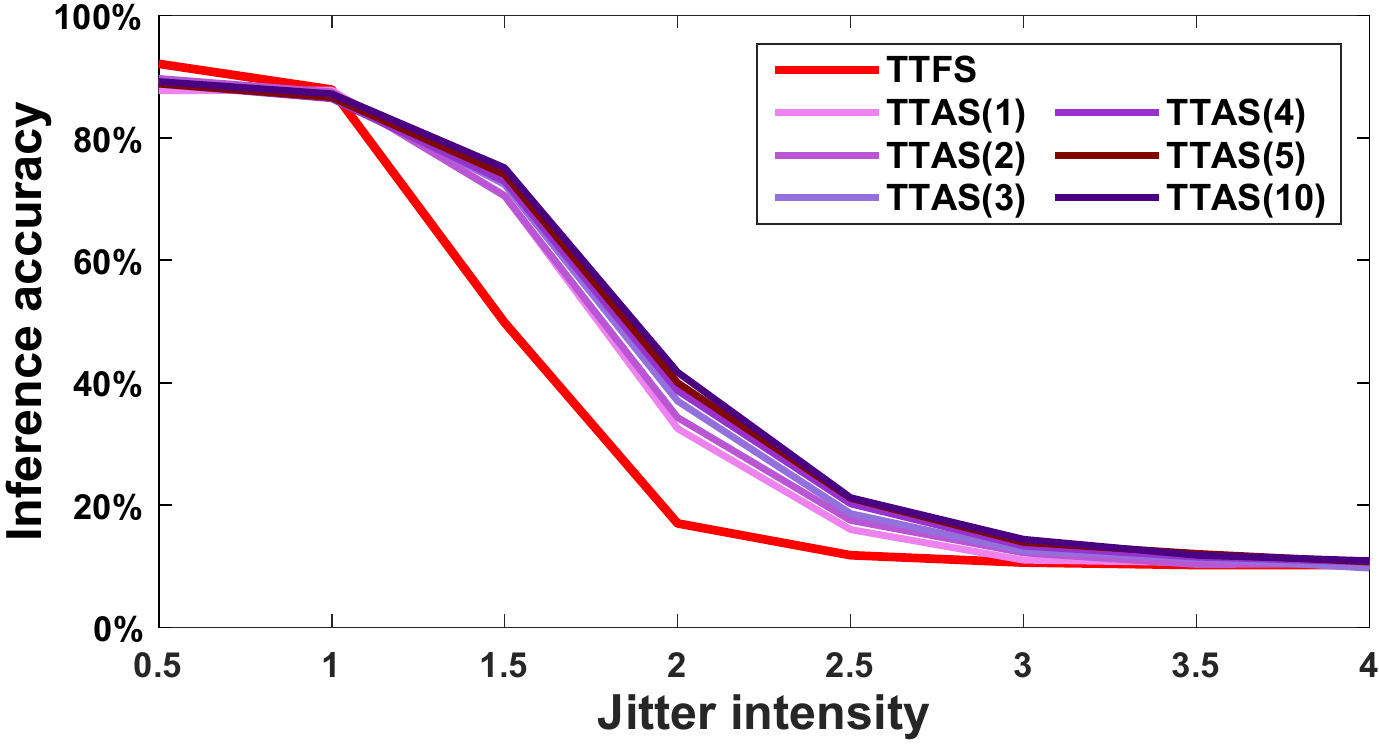}
	%\vspace{-0.5em}
	\caption{Inference accuracy of TTFS and TTAS on VGG16 and CIFAR-10 dataset according to jitter intensity}
	\label{fig:proposed_jit}
	%\vspace{-2.0em}
\end{figure}

\begin{figure}[t]
    \centering
    \includegraphics[width=1.0\linewidth]{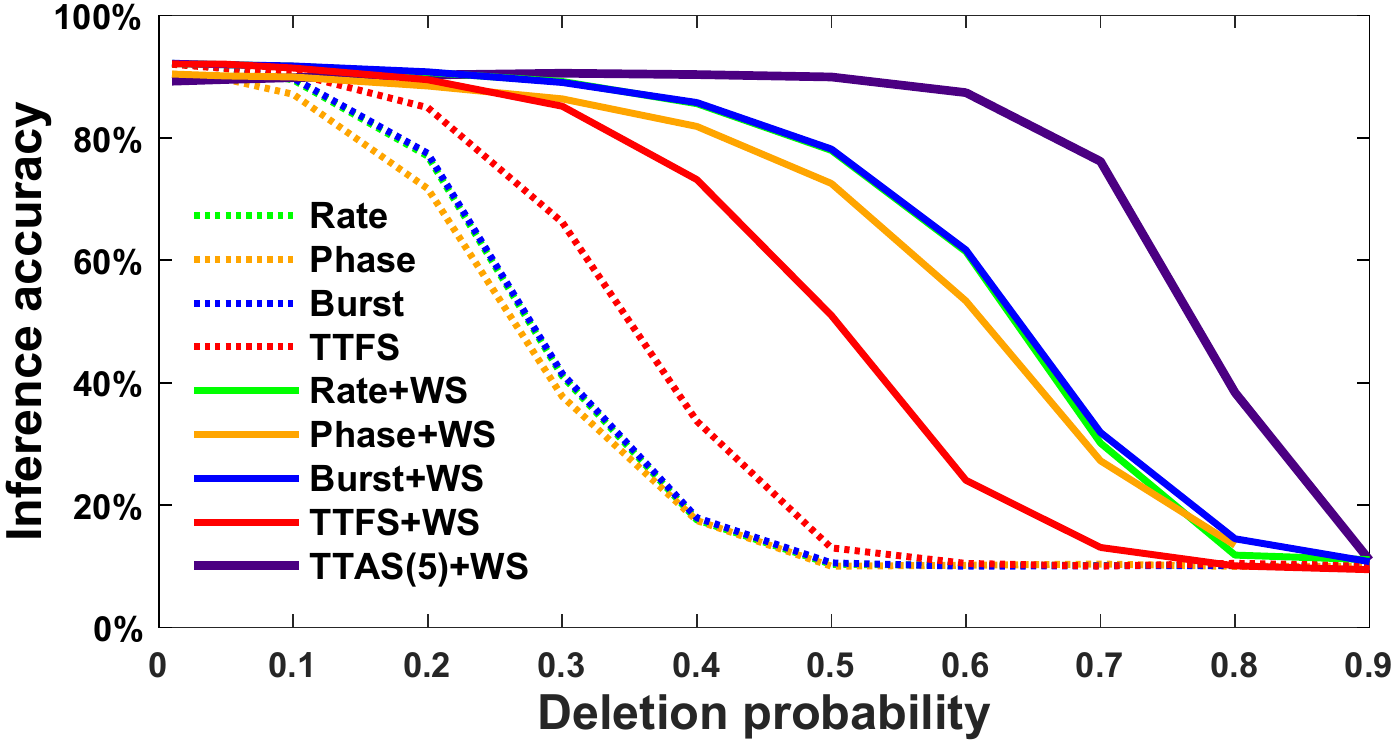}
	%\vspace{-0.5em}
	\caption{Comparisons of inference accuracy with various neural coding and proposed methods for spike deletion on VGG16 and CIFAR-10 dataset (WS: weight scaling)}
	\label{fig:exp_results_del}
	%\vspace{-2.0em}
\end{figure}

\begin{figure}[t]
    \centering
    \includegraphics[width=1.0\linewidth]{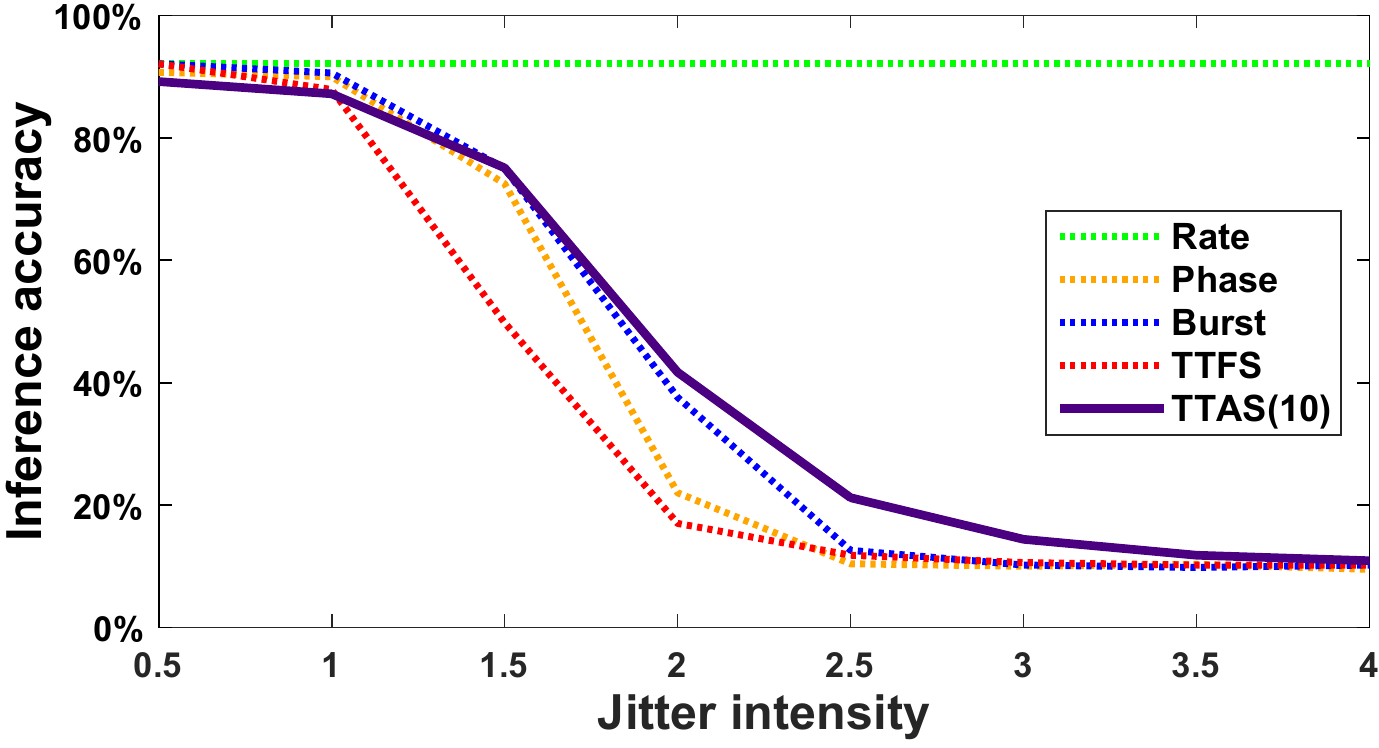}
	%\vspace{-0.5em}
	\caption{Comparisons of inference accuracy with various neural coding and proposed methods for spike jitter on VGG16 and CIFAR-10 dataset}
	\label{fig:exp_results_jit}
	\vspace{-1.0em}
\end{figure}

% setup
To evaluate the proposed methods, we experimented with various neural coding methods on MNIST, CIFAR-10, and CIFAR-100 datasets.
We empirically obtained the threshold $\theta$ to reduce inference latency and improve the efficiency of SNNs as in \cite{han2020rmp}.
As the threshold search results, we set $\theta$ to 0.4, 0.4, 1.2, and 0.8 for rate, burst, phase, and TTFS, respectively.
We used 1,000 time steps for CIFAR-10 and CIFAR-100 datasets and 100 time steps for MNIST dataset.
For TTFS, we set the time step to 108 on MNIST and CIFAR-10, and used 216 time steps on CIFAR-100.

The results of weight scaling and TTAS with deletion noise are depicted in Fig.~\ref{fig:proposed_del}.
With the proposed TTAS coding, the deep SNN showed robustness to the deletion noise as the target duration of burst spikes $t_{\textrm{a}}$ increased, which is represented in parentheses.
The results of TTAS on jitter noise are illustrated in Fig.~\ref{fig:proposed_jit}.
As the target duration increased, TTAS surpassed TTFS in terms of robustness.
In both noise cases of spike deletion and jitter, the improvements of TTAS were saturated as the target duration increased.

The comparisons with other neural coding and proposed methods are depicted in Fig.~\ref{fig:exp_results_del} and \ref{fig:exp_results_jit} for deletion and jitter noise, respectively.
We found the target duration of burst spikes empirically depending on the dataset and noise type.
As shown in Fig.~\ref{fig:exp_results_del}, by applying the weight scaling, the robustness against deletion was significantly improved.
However, when we compared the improvements between neural coding methods, TTFS showed the least improvement.
Our proposed approach, which is TTAS with weight scaling, was the most robust to the deletion noise.
For the jitter noise, TTAS could achieve similar robustness to that of burst coding.
The detailed experimental results, including the number of spikes, are reported in Tables~\ref{tab:experimental_result_del} and \ref{tab:experimental_result_jit}.

\setlength{\textfloatsep}{0pt}
\ctable[
pos = t,
star,
center,
caption = {Experimental results of spike deletion on deep SNNs with various neural coding methods according to deletion probability ($p$) (WS: weight scaling)},
%captionskip = -1.0ex,
%mincapwdth = \textwidth,
%width=\columnwidth,
label = {tab:experimental_result_del},
%doinside = {\small \def\arraystretch{.8}}
%doinside = {\footnotesize \def\arraystretch{.9} \setlength{\tabcolsep}{3pt}}
%doinside = {\small \def\arraystretch{.8} \setlength{\tabcolsep}{5pt}}
%doinside = {\small \def\arraystretch{1.0} \setlength{\tabcolsep}{4pt}}
doinside = {\small \def\arraystretch{1.0}}
]{l|c|cccc|c|cccc}{
    %@\tnote[a]{Efficiency (Eff.) = \# of Spikes per image / Neurons / Latency}
    %\tnote[a]{spiking density:= \# of spikes per image / (\# of neurons $\cdot$ latency)}
    %\tnote[a]{normalized energy estimation results for each task}
    %\tnote[c]{our experimental results}
    %\tnote[c]{CNN: 12c5-2s-64c5-2s-10}
    %\vspace{-6.4em}
}{
    \toprule
    %@\multirow{2}{*}{Methods} & \multicolumn{2}{c}{Neural Coding} & \multirow{2}{*}{Model} & Number of &\multicolumn{2}{c}{Accuracy (\%)} & \multirow{2}{*}{Latency} & \multicolumn{2}{c}{Spikes/image} \\
    %@& Input & Hidden & & Neurons & DNN & SNN & \quad & \# ($10^6$) & Eff. \\
    %Neural & Accuracy & \multirow{2}{*}{Latency} & Spikes & \multicolumn{2}{c}{Normalized Energy}\\
    %& \multicolumn{9}{c||}{Accuracy (\%)} & \multicolumn{9}{c}{Spikes}\\
    \multirow{2}{*}{Methods} & \multicolumn{5}{c|}{Accuracy (\%)} & \multicolumn{5}{c}{The number of spikes}\\
    & Clean & 0.2 & 0.5 & 0.8 & Avg. & Clean & 0.2 & 0.5 & 0.8 & Avg. \\
    %& Accuracy & \multicolumn{1}{c}{Latency} & \multicolumn{1}{c|}{Spikes} & \multicolumn{2}{c}{Normalized Energy}\\
    %& (\%) & \multicolumn{1}{c}{(time step)} & \multicolumn{1}{c|}{($10^6$)} & \multicolumn{1}{c}{TN~\cite{merolla2014million}} & \multicolumn{1}{c}{SN~\cite{furber2014spinnaker}} \\
	\midrule
	\multicolumn{6}{l}{MNIST} \\
	\midrule

	Rate + WS~\cite{han2020rmp} & 99.19 & 99.17 & 99.03 & \textbf{98.50} & 98.90 & 9.48E4 & 7.68E4 & 4.94E4 & 2.16E4 & 4.93E4 \\
	Phase + WS~\cite{kim2018deep} & 99.33 & 99.23 & \textbf{99.12} & 98.44 & \textbf{98.93} & 1.10E5 & 8.85E4 & 5.66E4 & 2.41E4 & 5.64E4 \\
	Burst + WS~\cite{park2019fast} & \textbf{99.34} & \textbf{99.29} & 98.19 & 98.79 & 99.09 & 9.06E4 & 7.31E4 & 4.64E4 & 1.95E4 & 4.64E4 \\
	%Reverse~\cite{zhang2019tdsnn} & 99.08 & - & - & -  & - \\
	TTFS + WS~\cite{park2020t2fsnn} & 99.31 & 99.20 & 98.26 & 64.90 & 87.45 &\textbf{3.05E3} & \textbf{2.54E3} & \textbf{1.72E3} & \textbf{7.65E2}& \textbf{1.68E3}\\
	\textbf{TTAS} + WS & \textbf{99.34} & 99.25 & 99.03 & 95.97 & 98.08 & 1.04E4 & 8.30E3 & 5.50E3 & 2.57E3 & 5.46E3 \\

	\midrule
	\multicolumn{6}{l}{CIFAR-10} \\
	\midrule
	Rate + WS~\cite{han2020rmp} & 92.15 & 90.68 & 77.95 & 11.82 & 60.15 & 1.71E7 & 1.30E7 & 8.72E6 & 4.70E6 & 8.78E6 \\
	Phase + WS~\cite{kim2018deep} & 90.55 & 88.48 & 72.55 & 13.44 & 58.16 & 2.22E7 & 1.73E7 & 1.12E7 & 5.80E6 & 1.14E7 \\
	Burst + WS~\cite{park2019fast} & \textbf{92.21} & \textbf{90.79} & 78.17 & 14.49 & 61.15 & 8.67E6 & 1.26E7 & 8.60E6 & 4.63E6 & 8.61E6\\
	TTFS + WS~\cite{park2020t2fsnn} & \textbf{92.21} & 89.50 & 50.99 & 10.06 & 49.65 & \textbf{9.18E4} & \textbf{7.98E4} & \textbf{5.99E4} & \textbf{3.12E4} & \textbf{5.66E4}\\
	\textbf{TTFA} + WS & 89.16 & 90.28 & \textbf{89.95} & \textbf{38.38} & \textbf{72.87} & 4.24E5 & 3.48E5 & 2.27E5 & 1.01E5 & 2.25E5 \\
	
	\midrule
	\multicolumn{6}{l}{CIFAR-100} \\
	\midrule
	Rate + WS~\cite{han2020rmp} & 67.39 & 64.37 & 39.95 & 1.67 & 35.33 & 2.00E7& 1.54E7 & 1.06E7 & 5.82E6 & 1.06E7 \\
	Phase + WS~\cite{kim2018deep} & 63.04 & 59.81 & 34.90 & \textbf{2.87} & 32.53 & 2.39E7 & 1.86E7 & 1.23E7 & 6.52E6 & 1.25E7\\
	Burst + WS~\cite{park2019fast} & 67.76 & \textbf{64.84} & 41.08 & 2.42 & 36.11 & 1.93E7 & 1.51E7 & 1.05E7 & 5.74E6 & 1.04E6 \\
	TTFS + WS~\cite{park2020t2fsnn} & \textbf{68.07} & 63.75 & 1.35 & 0.94 & 22.01 & \textbf{1.01E5} & \textbf{8.45E4} & \textbf{6.36E4} & \textbf{3.28E4} & \textbf{6.03E4}\\
	\textbf{TTFA + WS} & 64.63 & 61.24 & \textbf{54.26} & 2.42 & \textbf{39.31} & 1.79E5 & 1.48E5 & 9.93E4 & 4.54E4 & 9.75E4\\
	
	\bottomrule
}

\setlength{\textfloatsep}{0pt}
\ctable[
pos = t,
%star,
center,
caption = {Accuracy (\%) of spike jitter on deep SNNs with various neural coding methods depending on jitter intensity ($\sigma$)},
%captionskip = -1.0ex,
%mincapwdth = \textwidth,
%width=\columnwidth,
label = {tab:experimental_result_jit},
%doinside = {\small \def\arraystretch{.8}}
%doinside = {\footnotesize \def\arraystretch{.9} \setlength{\tabcolsep}{3pt}}
%doinside = {\small \def\arraystretch{.8} \setlength{\tabcolsep}{5pt}}
%doinside = {\small \def\arraystretch{1.0} \setlength{\tabcolsep}{4pt}}
doinside = {\small \def\arraystretch{1.0} }
]{l|c|cccc}{
    %@\tnote[a]{Efficiency (Eff.) = \# of Spikes per image / Neurons / Latency}
    %\tnote[a]{spiking density:= \# of spikes per image / (\# of neurons $\cdot$ latency)}
    %\tnote[a]{normalized energy estimation results for each task}
    %\tnote[c]{our experimental results}
    %\tnote[c]{CNN: 12c5-2s-64c5-2s-10}
    %\vspace{-6.4em}
}{
    \toprule
    %@\multirow{2}{*}{Methods} & \multicolumn{2}{c}{Neural Coding} & \multirow{2}{*}{Model} & Number of &\multicolumn{2}{c}{Accuracy (\%)} & \multirow{2}{*}{Latency} & \multicolumn{2}{c}{Spikes/image} \\
    %@& Input & Hidden & & Neurons & DNN & SNN & \quad & \# ($10^6$) & Eff. \\
    %Neural & Accuracy & \multirow{2}{*}{Latency} & Spikes & \multicolumn{2}{c}{Normalized Energy}\\
    %\multirow{2}{*}{Methods} & Clean & \multicolumn{4}{c|}{Accuracy (\%)} \\
    %Methods & Clean & \multicolumn{4}{c|}{Accuracy (\%)} \\
    Methods & Clean & 1.0 & 2.0 & 3.0 & Avg. \\
    %& Accuracy & \multicolumn{1}{c}{Latency} & \multicolumn{1}{c|}{Spikes} & \multicolumn{2}{c}{Normalized Energy}\\
    %& (\%) & \multicolumn{1}{c}{(time step)} & \multicolumn{1}{c|}{($10^6$)} & \multicolumn{1}{c}{TN~\cite{merolla2014million}} & \multicolumn{1}{c}{SN~\cite{furber2014spinnaker}} \\
	\midrule
	\multicolumn{6}{l}{MNIST} \\
	\midrule
	%Rate~\cite{diehl2015fast,rueckauer2017conversion} & 99.19 & 99.19 & 99.19 & 99.19 & 99.19 \\
	Phase~\cite{kim2018deep} & 99.33 & 99.17 & 93.20 & 56.36 & 82.91\\
	Burst~\cite{park2019fast} & \textbf{99.34} & \textbf{99.21} & 95.31 & 58.66 & 84.39 \\
	%Reverse~\cite{zhang2019tdsnn} & 99.08 & - & - & -  & - \\
	TTFS~\cite{park2020t2fsnn} & 99.31 & 99.20 & \textbf{98.76} & 60.58 & 83.45\\
	TTAS & \textbf{99.34} & 99.13 & 97.49 & \textbf{91.52} & \textbf{96.05} \\

	\midrule
	\multicolumn{6}{l}{CIFAR-10} \\
	\midrule
	%Rate~\cite{diehl2015fast,rueckauer2017conversion} & 91.14 & 92.15 & 92.15 & 92.15 & 92.15 \\
	Phase~\cite{kim2018deep} & 90.55 & 89.95 & 21.95 & 9.96 & 40.62 \\
	Burst~\cite{park2019fast} & \textbf{92.21} & \textbf{90.61} & 37.62 & 10.19 & 46.14 \\
	TTFS~\cite{park2020t2fsnn} & \textbf{92.21} & 87.86 & 16.99 & 10.57 & 38.47 \\
	TTAS & 89.16 & 86.58 & \textbf{40.01} & \textbf{13.91} & \textbf{46.83} \\
	
	\midrule
	\multicolumn{6}{l}{CIFAR-100} \\
	\midrule
	%Rate~\cite{diehl2015fast,rueckauer2017conversion} & 66.50 & 67.40 & 67.40 & 67.40 & 67.40 \\
	Phase~\cite{kim2018deep} & 63.04 & 59.12 & 1.54 & 1.01 & 20.56 \\
	Burst~\cite{park2019fast} & 67.76 & \textbf{63.78} & 2.06 & 1.23 & 22.36 \\
	TTFS~\cite{park2020t2fsnn} & \textbf{68.07} & 50.59 & 1.66 & 1.26 & 17.84 \\
	TTAS & 64.63 & 56.66 & \textbf{40.17} & \textbf{8.69} & \textbf{35.17} \\
	
	\bottomrule
}

%% file: 7-conclusion.tex
In this paper, we thoroughly analyzed the effect of spike noises, which are spike deletion and jitter, on deep SNNs.
Based on the analysis, we proposed a noise-robust deep SNN with weight scaling and TTAS coding.
The proposed SNN shows noise robustness while exploiting the efficiency of temporal coding without additional training procedures.
We believe that our approach paves the way to efficient and robust neuromorphic computing in the near future.

%% file: 8_acknowledgments.tex
This work was supported in part by the National Research Foundation of Korea (NRF) grant funded by the Korea government (Ministry of Science and ICT) [2016M3A7B4911115, 2018R1A2B3001628, 2021R1C1C2010454], the Brain Korea 21 Plus Project in 2021, AIRS Company in Hyundai Motor Company \& Kia Motors Corporation through HKMC-SNU AI Consortium Fund, and Samsung Research Funding \& Incubation Center of Samsung Electronics under Project Number SRFC-IT1901-12.

%SNN
%중견
%세종과학

%BK21

%현대컨소시움
%삼성미래재단

%This work was supported by AIRS Company in Hyundai Motor Company \& Kia Motors Corporation through HKMC-SNU AI Consortium Fund.
%Samsung Research Funding \& Incubation Center of Samsung Electronics under Project Number SRFC-IT1901-12

%This work was supported in part by the National Research Foundation of Korea (NRF) grant funded by the Korea government (Ministry of Science and ICT) [2016M3A7B4911115, 2018R1A2B3001628], the Brain Korea 21 Plus Project in 2019, AIR Lab (AI Research Lab) in Hyundai Motor Company through HMC-SNU AI Consortium Fund, and Samsung Research Funding \& Incubation Center of Samsung Electronics under Project Number SRFC-IT1901-12.

%% file: 0-0_main.bbl
\begin{thebibliography}{10}
\providecommand{\url}[1]{#1}
\csname url@samestyle\endcsname
\providecommand{\newblock}{\relax}
\providecommand{\bibinfo}[2]{#2}
\providecommand{\BIBentrySTDinterwordspacing}{\spaceskip=0pt\relax}
\providecommand{\BIBentryALTinterwordstretchfactor}{4}
\providecommand{\BIBentryALTinterwordspacing}{\spaceskip=\fontdimen2\font plus
\BIBentryALTinterwordstretchfactor\fontdimen3\font minus
  \fontdimen4\font\relax}
\providecommand{\BIBforeignlanguage}[2]{{%
\expandafter\ifx\csname l@#1\endcsname\relax
\typeout{** WARNING: IEEEtran.bst: No hyphenation pattern has been}%
\typeout{** loaded for the language `#1'. Using the pattern for}%
\typeout{** the default language instead.}%
\else
\language=\csname l@#1\endcsname
\fi
#2}}
\providecommand{\BIBdecl}{\relax}
\BIBdecl

\bibitem{park2019fast}
S.~Park, S.~Kim, H.~Choe, and S.~Yoon, ``Fast and efficient information
  transmission with burst spikes in deep spiking neural networks,'' in
  \emph{Design Automation Conference (DAC)}, 2019.

\bibitem{kim2020spiking}
S.~Kim, S.~Park, B.~Na, and S.~Yoon, ``Spiking-yolo: spiking neural network for
  energy-efficient object detection,'' in \emph{AAAI Conference on Artificial
  Intelligence (AAAI)}, 2020.

\bibitem{park2020t2fsnn}
S.~Park, S.~Kim, B.~Na, and S.~Yoon, ``T2fsnn: deep spiking neural networks
  with time-to-first-spike coding,'' in \emph{Design Automation Conference
  (DAC)}, 2020.

\bibitem{gautrais1998rate}
J.~Gautrais and S.~Thorpe, ``Rate coding versus temporal order coding: a
  theoretical approach,'' \emph{Biosystems}, vol.~48, no. 1-3, pp. 57--65,
  1998.

\bibitem{kim2018deep}
J.~Kim, H.~Kim, S.~Huh, J.~Lee, and K.~Choi, ``Deep neural networks with
  weighted spikes,'' \emph{Neurocomputing}, vol. 311, pp. 373--386, 2018.

\bibitem{merolla2014million}
P.~Merolla \emph{et~al.}, ``A million spiking-neuron integrated circuit with a
  scalable communication network and interface,'' \emph{Science}, vol. 345, no.
  6197, pp. 668--673, 2014.

\bibitem{davies2018loihi}
M.~Davies \emph{et~al.}, ``Loihi: A neuromorphic manycore processor with
  on-chip learning,'' \emph{IEEE Micro}, vol.~38, no.~1, pp. 82--99, 2018.

\bibitem{roy2019towards}
K.~Roy, A.~Jaiswal, and P.~Panda, ``Towards spike-based machine intelligence
  with neuromorphic computing,'' \emph{Nature}, vol. 575, no. 7784, pp.
  607--617, 2019.

\bibitem{bouvier2019spiking}
M.~Bouvier \emph{et~al.}, ``Spiking neural networks hardware implementations
  and challenges: A survey,'' \emph{ACM Journal on Emerging Technologies in
  Computing Systems}, vol.~15, no.~2, pp. 1--35, 2019.

\bibitem{querlioz2011simulation}
D.~Querlioz, O.~Bichler, and C.~Gamrat, ``Simulation of a memristor-based
  spiking neural network immune to device variations,'' in \emph{International
  Joint Conference on Neural Networks (IJCNN)}, 2011.

\bibitem{querlioz2013immunity}
D.~Querlioz, O.~Bichler, P.~Dollfus, and C.~Gamrat, ``Immunity to device
  variations in a spiking neural network with memristive nanodevices,''
  \emph{IEEE Transactions on Nanotechnology}, vol.~12, no.~3, pp. 288--295,
  2013.

\bibitem{cheng2020lisnn}
X.~Cheng, Y.~Hao, J.~Xu, and B.~Xu, ``Lisnn: Improving spiking neural networks
  with lateral interactions for robust object recognition,'' in
  \emph{International Joint Conference on Artificial Intelligence (IJCAI)},
  2020.

\bibitem{yu2015spiking}
Q.~Yu, R.~Yan, H.~Tang, K.~C. Tan, and H.~Li, ``A spiking neural network system
  for robust sequence recognition,'' \emph{IEEE transactions on neural networks
  and learning systems}, vol.~27, no.~3, pp. 621--635, 2015.

\bibitem{zhang2019fast}
A.~Zhang, H.~Zhou, X.~Li, and W.~Zhu, ``Fast and robust learning in spiking
  feed-forward neural networks based on intrinsic plasticity mechanism,''
  \emph{Neurocomputing}, vol. 365, pp. 102--112, 2019.

\bibitem{zheng2018sparse}
Y.~Zheng, S.~Li, R.~Yan, H.~Tang, and K.~C. Tan, ``Sparse temporal encoding of
  visual features for robust object recognition by spiking neurons,''
  \emph{IEEE transactions on neural networks and learning systems}, vol.~29,
  no.~12, pp. 5823--5833, 2018.

\bibitem{chowdhury2020towards}
S.~S. Chowdhury, C.~Lee, and K.~Roy, ``Towards understanding the effect of leak
  in spiking neural networks,'' \emph{arXiv preprint arXiv:2006.08761}, 2020.

\bibitem{izhikevich2004model}
E.~M. Izhikevich, ``Which model to use for cortical spiking neurons?''
  \emph{IEEE transactions on neural networks}, vol.~15, no.~5, pp. 1063--1070,
  2004.

\bibitem{maass1997networks}
W.~Maass, ``Networks of spiking neurons: the third generation of neural network
  models,'' \emph{Neural Networks}, vol.~10, no.~9, pp. 1659--1671, 1997.

\bibitem{gerstner2014neuronal}
W.~Gerstner, W.~M. Kistler, R.~Naud, and L.~Paninski, \emph{Neuronal dynamics:
  From single neurons to networks and models of cognition}.\hskip 1em plus
  0.5em minus 0.4em\relax Cambridge University Press, 2014.

\bibitem{zhang2020temporal}
W.~Zhang and P.~Li, ``Temporal spike sequence learning via backpropagation for
  deep spiking neural networks,'' in \emph{Conference on Neural Information
  Processing Systems (NeurIPS)}, 2020.

\bibitem{rueckauer2018conversion}
B.~Rueckauer and S.-C. Liu, ``Conversion of analog to spiking neural networks
  using sparse temporal coding,'' in \emph{International Symposium on Circuits
  and Systems (ISCAS)}, 2018.

\bibitem{han2020rmp}
B.~Han, G.~Srinivasan, and K.~Roy, ``Rmp-snn: Residual membrane potential
  neuron for enabling deeper high-accuracy and low-latency spiking neural
  network,'' in \emph{Conference on Computer Vision and Pattern Recognition
  (CVPR)}, 2020.

\bibitem{hendrycks2019benchmarking}
D.~Hendrycks and T.~Dietterich, ``Benchmarking neural network robustness to
  common corruptions and perturbations,'' in \emph{International Conference on
  Learning Representations (ICLR)}, 2019.

\bibitem{wunderlich2019demonstrating}
T.~Wunderlich \emph{et~al.}, ``Demonstrating advantages of neuromorphic
  computation: a pilot study,'' \emph{Frontiers in neuroscience}, vol.~13, p.
  260, 2019.

\bibitem{stromatias2015robustness}
E.~Stromatias, D.~Neil, M.~Pfeiffer, F.~Galluppi, S.~B. Furber, and S.-C. Liu,
  ``Robustness of spiking deep belief networks to noise and reduced bit
  precision of neuro-inspired hardware platforms,'' \emph{Frontiers in
  neuroscience}, vol.~9, p. 222, 2015.

\bibitem{li2020robustness}
C.~Li, R.~Chen, C.~Moutafis, and S.~Furber, ``Robustness to noisy synaptic
  weights in spiking neural networks,'' in \emph{International Joint Conference
  on Neural Networks (IJCNN)}, 2020.

\bibitem{neftci2016stochastic}
E.~O. Neftci, B.~U. Pedroni, S.~Joshi, M.~Al-Shedivat, and G.~Cauwenberghs,
  ``Stochastic synapses enable efficient brain-inspired learning machines,''
  \emph{Frontiers in neuroscience}, vol.~10, p. 241, 2016.

\bibitem{yu2020synaptic}
Q.~Yu, S.~Song, C.~Ma, L.~Pan, and K.~C. Tan, ``Synaptic learning with
  augmented spikes,'' \emph{arXiv}, 2020.

\bibitem{wu2018spiking}
J.~Wu, Y.~Chua, M.~Zhang, H.~Li, and K.~C. Tan, ``A spiking neural network
  framework for robust sound classification,'' \emph{Frontiers in
  neuroscience}, vol.~12, p. 836, 2018.

\bibitem{yu2020robust}
Q.~Yu, Y.~Yao, L.~Wang, H.~Tang, J.~Dang, and K.~C. Tan, ``Robust environmental
  sound recognition with sparse key-point encoding and efficient multispike
  learning,'' \emph{IEEE Transactions on Neural Networks and Learning Systems},
  2020.

\bibitem{srivastava2014dropout}
N.~Srivastava, G.~Hinton, A.~Krizhevsky, I.~Sutskever, and R.~Salakhutdinov,
  ``Dropout: a simple way to prevent neural networks from overfitting,''
  \emph{The journal of machine learning research}, vol.~15, no.~1, pp.
  1929--1958, 2014.

\bibitem{wilson1999simplified}
H.~R. Wilson, ``Simplified dynamics of human and mammalian neocortical
  neurons,'' \emph{Journal of theoretical biology}, vol. 200, no.~4, pp.
  375--388, 1999.

\end{thebibliography}
